# DeepFEA: Deep Learning for Prediction of Transient Finite Element Analysis Solutions

Georgios Triantafyllou[a], Panagiotis G. Kalozoumis[a], George Dimas[a],
and Dimitris K. Iakovidis[a,*]

[a]Dept. Computer Science and Biomedical Informatics, University of Thessaly, 2-4 Papasiopoulou st. 35131, Lamia, Greece

{gtriantafyllou, pkalozoumis, gdimas, diakovidis}@uth.gr

**Abstract**: Finite Element Analysis (FEA) enables the simulation of physical phenomena under various conditions. This is usually a computationally expensive and time-consuming process since it involves the numerical solution of partial differential equations. To accelerate FEA, effective surrogate methods based on Artificial Neural Networks (ANNs) have been recently proposed. However, these methods still have several limitations, mainly with respect to the dynamic prediction of accurate solutions independently from the original finite element model and/or respective ground truth data. This study proposes a deep learning-based framework for FEA (DeepFEA) that copes with these limitations. It is based on a novel ANN architecture composed of a multilayer Convolutional Long Short-Term Memory (ConvLSTM) network branching into two parallel convolutional neural networks with tensor outputs used to infer predictions related to the nodes and elements of FEA models. The architecture of the proposed network is optimized using a novel adaptive learning algorithm, called Node-Element Loss Optimization (NELO), which minimizes the error occurring at both of its branches. DeepFEA relies only on the initial and boundary conditions, as well as the external load of the modeled structure, which are provided as input to predict the transient solutions of an entire FEA simulation. The experimental evaluation of DeepFEA is performed on three datasets in the context of structural mechanics, generated to serve as publicly available reference datasets. The results indicate that it can accurately predict the outcome of multi-timestep FEA simulations, while offering a significant solution acceleration of at least two orders of magnitude.

## 1 Introduction

Physical phenomena can often be described by a system of equations, such as Partial Differential Equations (PDEs) or Ordinary Differential Equations (ODEs). PDEs are equations containing partial derivatives that encapsulate physical quantities of natural processes. The solution of these equations is provided by a function, the exact solution of which cannot be analytically calculated or is not unique. In research fields, such as mechanical and biomedical engineering, the behavior of objects that are subject to a system of PDEs is simulated via computational modeling tools that are able to solve these PDEs in a computationally efficient way. Finite Element Analysis (FEA) is often used to divide such objects into a finite number of discrete elements and nodes, enabling dedicated FEA software to solve the PDEs in a more efficient manner [1].

FEA-based simulations rely on iteratively solving PDEs that correspond to specific physical models. Usually, the duration of the problem to be simulated needs to be discretized into a number of timesteps; a large number of timesteps results in a solution of a higher resolution in the time domain and it usually entails a higher computational cost. At each timestep, FEA relies on the predictions of the previous state to predict the solution for the current state. These simulations can also become computationally expensive as the order of the PDEs and the density of the mesh increase. Reduced-order modelling (ROM) and surrogate models have been considered to address this problem. ROM solutions are mostly employed to reduce the dimensionality of high-order models [2–4], while surrogate models are used to replace FEA. Traditional ROM methods, *e.g.,* the proper orthogonal decomposition method and the principal component analysis method [5], are constrained by their linear nature; thus, AI-based solutions, such as Artificial Neural Networks (ANNs) that include Multilayer Perceptrons

---

(MLPs) and Convolutional Neural Networks (CNNs), are preferred for reducing the dimensionality of non-linear physical models [6, 7]. There has been an increasing interest in AI-based surrogate models that utilize ANNs [2], which is due to their inherent ability to approximate linear and non-linear functions that are embedded in their weights [8]. ANN architectures, such as MLPs [7], CNNs [9], Long Short-Term Memory (LSTM) [10], Gated Recurrent Unit (GRU) [11], Graph Neural Networks (GNNs) [12], and Physics-Informed Neural Networks (PINNS) [13], have been used to predict the solutions of FEA-based simulations [3]. Based on their application, surrogate models can be divided into three main categories, namely, steady-state, physics-informed, and transient analysis models.

Surrogate models that are developed to predict the solutions of steady-state simulations utilize only the input and initial conditions provided at a certain time-point without accounting for previous states or iteratively predicting the intermediate states that are necessary in traditional FEA. Recent surrogate models have been mainly based on MLPs and CNNs to predict the solutions of steady-state FEA simulations. Some of the MLP-based approaches predict parts of the solution assisted by FEA [14, 15]. In later studies, FEA was completely bypassed and MLPs were utilized to incorporate the FEA-related parameters in the input feature vectors [4, 6, 7, 16, 17]. Although this approach can be simple to implement, it can become computationally expensive for large meshes. A drawback of MLPs is that they lack spatial awareness, in the sense that they do not embed information related to the spatial location from which the features are extracted. Studies have attempted to address this problem by incorporating spatial information in the input vector, *e.g.*, geometry-related parameters [16–18]. CNNs inherently offer spatial awareness and they have been applied for predicting the outcome of FEA simulations, *e.g.*, deformation, stress, and strain [3, 9, 19, 20]. CNNs can scale better when considering larger meshes that require deeper networks. Bayesian CNNs have also been explored as surrogate models, providing adaptability in cases where a FEA simulation involves external loads [21, 22]. Notably, these methods utilize CNN architectures that are similar to U-Nets. Other CNN-based applications include multi-grid CNNs [23] and image coloring techniques for 2D microstructure-based FEA simulations, where the predictions are based on the pixel intensity of the microstructure in grayscale [24]. CNNs are suitable for structured meshes, which limits their capabilities in scenarios with high geometric complexity. To address this problem, the use of zero-padding has been proposed, where zero values are added to form an evenly structured tensor [20–22]. Despite their potential for steady-state predictions, these methods lack temporal awareness; thus, they are not suitable for modeling scenarios where all the states of a simulation need to be predicted. Since the next state of the mesh (deformation) depends on the previous timesteps, *i.e.,* the node coordinates affect the predicted deformation in the next timestep, the error propagates from the previous to the current timestep, leading to large errors toward the end of the simulation.

Physics-informed methods have also been explored in the context of approximating FEA simulations [13, 25–27]. These methods have recently gained popularity due to their ability to incorporate the PDE in the loss function by utilizing the automatic differentiation process of ANNs achieved via back propagation. The most popular category are PINNs, where the input contains spatial variables, *i.e.*, coordinates of the mesh nodes, and a time variable [13, 25, 27, 28]. Implicit neural representations (INRs) have also been utilized in a similar way as PINNs, where the boundary conditions are included in the loss function by incorporating spatiotemporal information based on the gradients of previous timesteps in the loss function [26]. These approaches are excellent choices for interpolation, since they can learn to generate results over the whole domain of the experiment based only on few training data. However, a major limitation is that they lack generalizability, in the sense that the loss function used for training such networks needs to be different for different applications and boundary conditions; thus, once trained, their application is limited to specific scenarios with invariable simulation characteristics.

Transient analysis prediction methodologies consider the temporal aspects of physical problems, and they can be utilized to predict the solution for all the intermediate states; thus, they can overcome the problems arising from steady-state prediction approaches. These methods mainly consider the prediction of one or more states based on previous timesteps. In this context, MLPs with the assistance of non-linear autoregressive exogenous models (NARXs) have been explored [29, 30]. Nonetheless, they require a substantial number of timesteps to initialize the NARX model and are simulation-specific. Recurrent neural networks (RNNs) embed the temporal information of time-series in their weights, as well as in dedicated variables that contain information about the output of the previous timesteps. State-of-the-art methods that utilize RNNs, such as LSTM or GRU, have been used as surrogate models for PDE-based simulations [31], but they can partially predict parts of the solution based on certain calculated output parameters provided by FEA [10, 11, 32]. It should be noted that, in such simulations, the input of the current timestep is not affected by the output of the previous timestep and there is no need for a spatially-aware methodology. Therefore, the expected output can be easily predicted based only on the time-coupling that RNNs provide.

FEA simulations that depend on spatiotemporal correlations between input conditions and predicted output parameters can benefit from the combination of CNNs with RNNs. The fusion of CNNs and LSTMs (CNN-LSTMs) has been adopted for such tasks [33–36]. Nevertheless, these methods are limited to 2D scenarios, where only one output parameter of the solution is predicted. CNN-LSTMs have also been applied in Fluid-Structure Interaction (FSI) simulations, where the output of the structural solution is used as input for the fluid solution and *vice versa* [33–35]. Graph Neural Networks (GNNs) have been recently used to predict FEA simulation solutions, since they have a similar spatial awareness capability to CNNs. GNNs can be combined with Auto Encoders (AEs), CNNs, GRUs, and LSTMs to create spatiotemporal-aware GNNs [12, 37]. Despite their benefits, these methods do not consider the dynamic nature of the input parameters and they are limited to inferring solutions only for a few timesteps. In addition, GNNs are limited in terms of scalability and can become computationally expensive for graphs with dynamic characteristics, which is a key feature of transient FEA simulations [38]. Convolutional LSTMs (ConvLSTMs) are another type of CNN and LSTM fusion, which have been used as surrogate models of FEA simulations. Methods that utilize ConvLSTMs have been applied to microstructure-based FEA simulations in the 2D domain [39], material design [40], and surface heat-flux distribution [41]. ConvLSTMs can extract spatiotemporal information from a time-series of images, such as videos, without requiring complex ANN or RNN architectures, thereby providing more scalability and flexibility. This flexibility can be particularly advantageous when dealing with FEA simulations, which often involve diverse mesh structures and varying levels of complexity [41]. However, in the context of FEA simulations, they are often used for forecasting the evolution of output parameters based on an initial amount of ground truth data [39, 40].

Although various surrogate models have been proposed to predict FEA solutions, current models are limited to steady-state predictions, they cannot completely replace FEA, they have limited generalizability, and/or rely on an initial amount of ground truth data. Furthermore, recurrent models, which rely on previous predictions to provide future solution estimations, usually result in error propagation and accumulation over time. To cope with all these limitations, this study proposes a novel, generic deep learning framework called DeepFEA, which is capable of completely replacing FEA, providing accurate and time-efficient predictions of transient FEA solutions. More specifically, the contributions of this study include:

- A novel deep ANN architecture composed of a multilayer ConvLSTM and two parallel CNN branches. Unlike relevant state-of-the-art architectures, it considers as input the coordinates of the mesh nodes, the initial and boundary conditions, and the external load, to dynamically predict the solutions of 2D and 3D transient FEA simulations. The predictions for the nodes and the elements of the FEA model are made separately in the form of tensors.

- A novel adaptive learning algorithm, called Node-Element Loss Optimization (NELO), inspired from a technique used in the context of Natural Language Processing (NLP), called Scheduled Sampling Method (SSM) [42]. The SSM algorithm has been adapted by utilizing a novel loss function co-considering the node- and element-level errors to enable a progressive incorporation of the previous predictions to each subsequent timestep, and the minimization of the accumulated error over several timesteps.

- A total of three, publicly available, reference datasets generated by 2D and 3D FEA model simulations in the context of structural mechanics, enabling comparisons with surrogate models.
- An experimental study that uses these datasets to assess the proposed framework both quantitatively and qualitatively.

The remainder of this paper is organized as follows: Section 2 describes the proposed framework; Section 3 presents its evaluation process for different 2D and 3D cases; Section 4 discusses the results and the key findings, and Section 5 summarizes the conclusions of this study and future work.

## 2 Methods

DeepFEA is a deep learning framework, which, given a set of initialization parameters of a FEA simulation, can be used to simultaneously predict node- and element-related output parameters of the FEA after multiple timesteps. The core component of DeepFEA is a novel network architecture, which is trainable. The network can provide very accurate predictions after being trained with NELO, which is a novel training algorithm especially designed to minimize the error in transient FEA simulations. An overview of the proposed framework, comprising three stages, namely the dataset generation phase (Fig. 1a), training phase (Fig. 1b) and testing phase (Fig. 1c), is presented in Fig. 1.

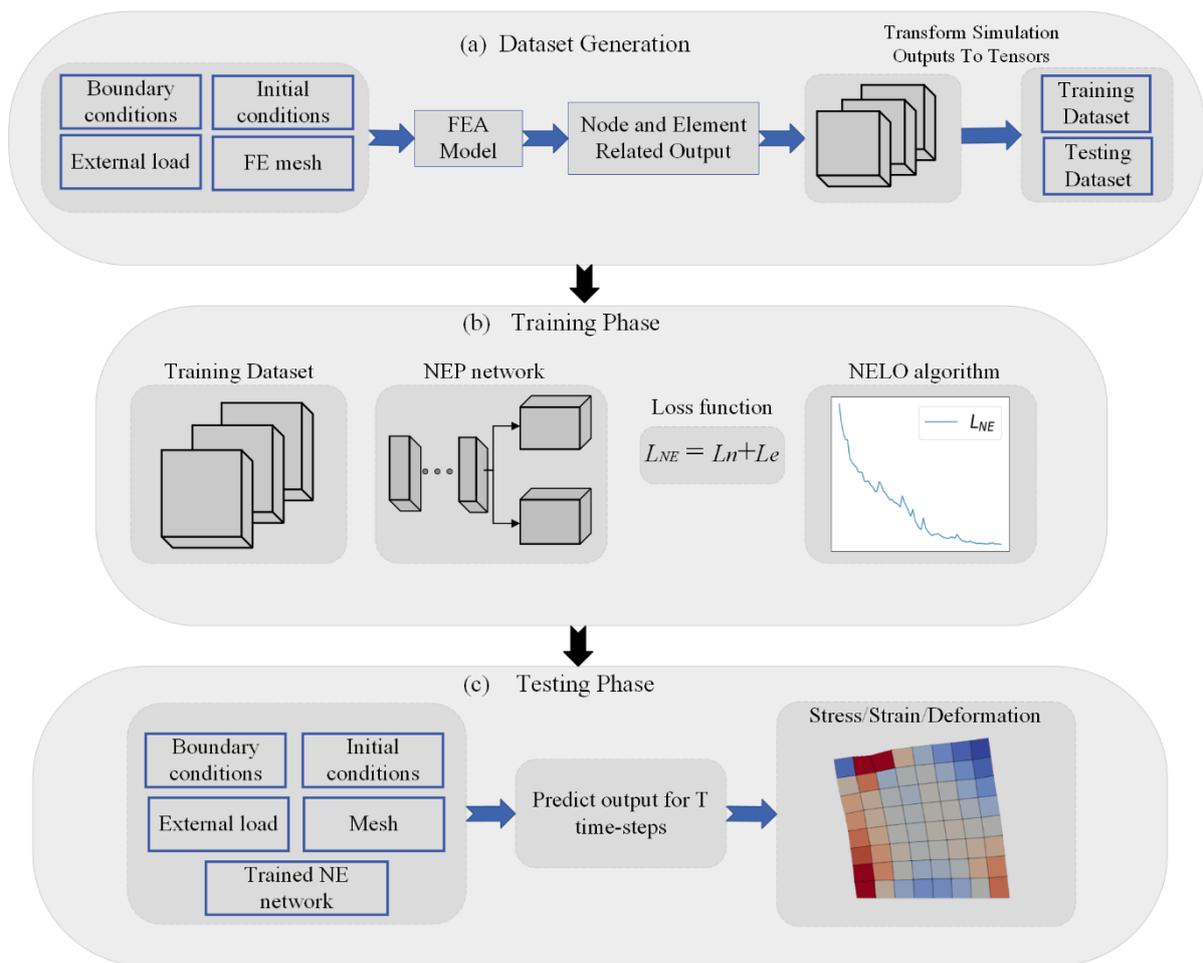

*Figure 1. DeepFEA framework.*

The first phase of the DeepFEA framework (Fig. 1a) involves the generation of a transient FEA simulation dataset. Boundary and initial conditions, mesh structure (FE mesh), and varying maximum external loads are selected to create a dataset with representative transient FEA simulations that will enable the network to learn the behavior of the examined FEA model. Then, a commercial FEA simulation software is used to calculate the solutions for each simulation and a pre-processing step maps the node- and element-related outputs of every timestep to tensors that will be used as input and output for the network. Subsequently, (Fig. 1b) a portion of the dataset is used to train the network based on the NELO algorithm (Subsection 2.2.1) which aims to minimize the loss function $\mathcal{L}_{NE}$ (Subsection 2.2.2).

Once the network is trained (Fig. 1c), it can predict the solutions of simulations on the same mesh structure that have not been included in the construction of the training dataset. This approach allows for recurrent estimation of the simulation output in the time domain. In the following subsections the details of the architecture and of the training phase are described.

## 2.1 Node-Element-based Network Architecture

The Node-Element Prediction (NEP) network architecture of DeepFEA consists of two modules, a Feature Extraction Module (FExM) and a Prediction Module (PM). FExM is implemented using a multilayer ConvLSTM network, which receives as input the FEA model characteristics and the initialization parameters for the FEA in the form of a tensor. FExM is tasked to automatically extract spatiotemporal features in the form of multiple feature maps extracted over consecutive timesteps. More specifically, the feature maps encode spatial information regarding the nodes and elements of the FEA model along with the respective physical properties considered in the FEA simulation. This information is propagated through each layer of the ConvLSTM network, allowing it to progressively extract complex spatial and physics-related information of the simulation in various degrees of abstraction. The output of the final ConvLSTM layer encapsulates the overall spatial structure of the mesh, the information retained from previous timesteps, as well as extracted information about the physics of the simulation, serving as a high-level representation of the FEA simulation. Given a set of extracted features by FExM, PM predicts the output parameters of the simulation. PM comprises two parallel CNN branches, producing two output tensors that follow the structure of the input tensor. These output tensors have different dimensions corresponding to the node- and element-related parts of the FE mesh.

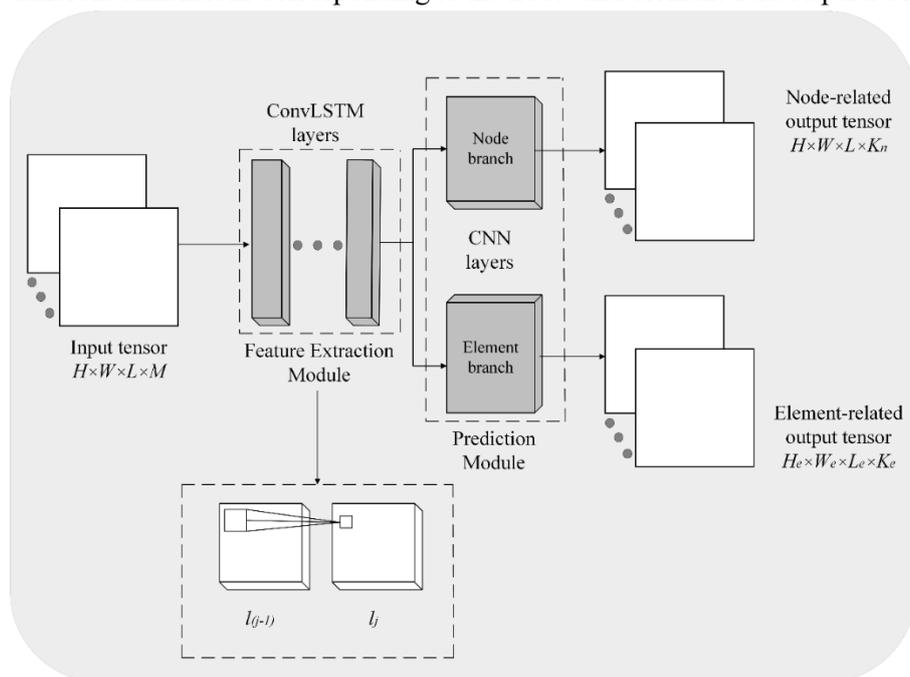

*Figure 2. The DeepFEA NEP network architecture.*

### 2.1.1    Input and Output Tensors

The geometry of the structure to be analyzed with FEA is first discretized into a digital representation, called FE mesh, which is composed of a finite number of nodes (discretized points) and elements (area or volume that connects the nodes of the mesh). Each node and element of the mesh stores information related to the FEA model, *e.g.*, coordinates or stress/strain in each node or element, respectively. The FEA model is created by specifying a mesh structure, the initial conditions, the boundary conditions, and the externally applied load. In this study, the coordinates of the mesh are considered as the initial conditions, the positions of the constrained nodes as the boundary conditions, and the forces applied on the outer nodes as the externally applied load. A FEA simulation is tasked to predict the output parameters for each node and element of the mesh based on the initial conditions corresponding to each node, external load, and boundary conditions.

In Fig. 2, the input tensor size is $H \times W \times L$, where $H$, $W$, and $L$ denote the height, width, and length of each input feature map, respectively, whereas $M$ denotes the number of the different parameters considered for solving a specific problem. For a 3D mesh, $H \times W \times L$ correspond to the dimensions of the input mesh. The notation $H_e \times W_e \times L_e$ refers to the dimensions of a tensor that contains information regarding the elements of the mesh. In the 2D mesh case, only the height and the width of the mesh are considered, *i.e.*, $H$ and $W$. The output tensor of the node branch is of shape $H \times W \times K_n$ or $H \times W \times L \times K_n$ for 2D and 3D models, respectively. In this paper, the 2D and 3D FEA models are meshed with quadrilateral and hexahedral elements, respectively. Hence, the output tensor of the element branch is of shape $H_e \times W_e \times K_e$ for 2D or $H_e \times W_e \times L_e \times K_e$ for 3D FEA models, with $H_e = H$-1, $W_e = W$-1, and $L_e = L$-1. Variables $K_n$ and $K_e$ denote the number of different output targets estimated by the dedicated node and element branches, respectively, *e.g.*, displacement across the *x*-axis, displacement across the *y*-axis, *etc.* For example, considering the input referring to the position of each node for a 2D FE mesh, *i.e.*, a pair of *x* and *y* coordinates, the information of the *x* and *y* coordinates is structured into two different matrices, as illustrated in Fig. 3. In contrast to traditional CNN-based frameworks that employ padding techniques to accommodate structural changes in the input tensor, DeepFEA utilizes the FE mesh coordinates as input features. This approach enables the prediction of FEA solutions whilst the structure of the mesh changes due to deformation.

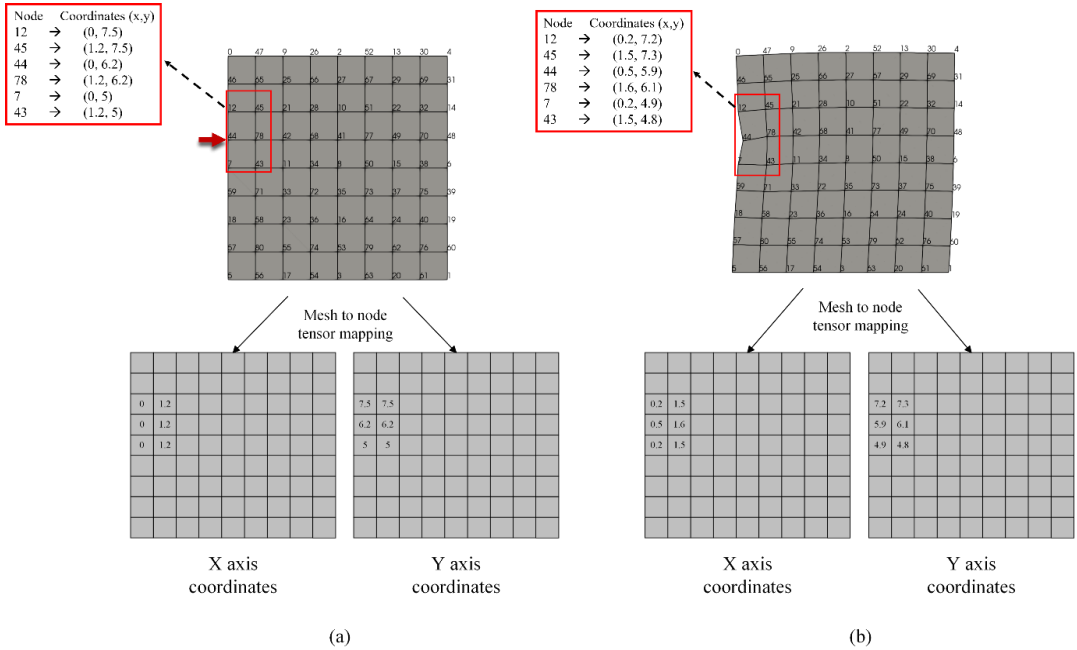

*Figure 3. Mapping of a 2D FE mesh to input and output node tensors. (a) Before and (b) after the application of a load. The red arrow indicates the position and direction of the applied force.*

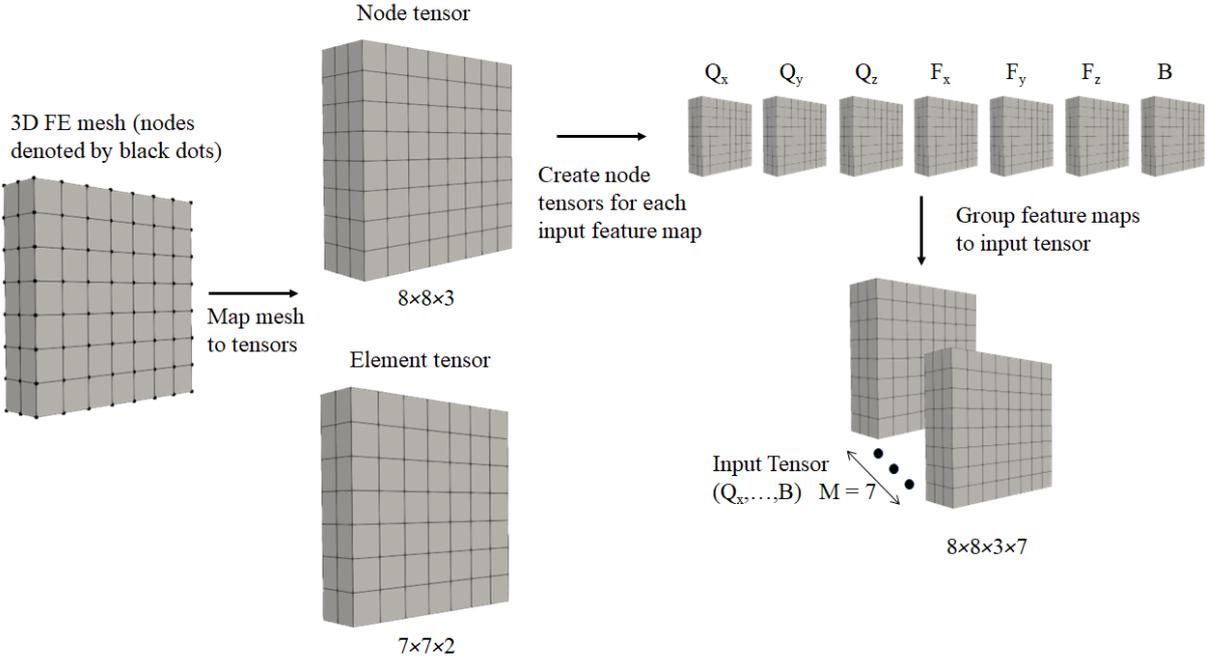

*Figure 4. Mapping process of a 3D FE mesh into node and element tensors..*

An example of how the input tensor of DeepFEA is formed for a 3D FEA model is presented in Fig. 4. The input tensor $X_t = (Q_x, Q_y, Q_z, F_x, F_y, F_z, B)$ consists of 7 input feature maps, where $(Q_x, Q_y, Q_z)$ is a triplet of feature maps denoting the coordinates of the mesh, $(F_x, F_y, F_z)$ is a triplet of feature maps denoting the initial force vectors applied to the mesh, and $B$ is a feature map denoting the constrained nodes for the timestep $t$. In this case, the dimensions of an input tensor $X_t$ are $8×8×3×7$, where $8×8×3$ are the dimensions of the input mesh and $7$ is the number of input features. Thus, each input feature map of $X_t$ can be considered as a copy of the initial node tensor with shape $8×8×3$, containing feature-related information. Furthermore, the fixed nodes in the feature map $B$ are denoted with 0, whereas non-constrained nodes are denoted with 1. This feature essentially works as a flag that minimizes the effect of the fixed nodes inside the feedforward process of the NEP network and enables it to differentiate between constrained and non-constrained nodes. Regarding the forces applied on the nodes of the mesh, at each timestep $t$, only the external load is considered as input. Hence, the $(F_x, F_y, F_z)$ channels contain information regarding only the forces applied externally during the simulation.

The output parameters considered in this study are the displaced coordinates of the mesh and the effective stress and strain for the node and element branches, respectively. Therefore, the output tensor of the node branch for a 3D FEA model can be defined as $Y_{t+1}^n = (Q'_x, Q'_y, Q'_z)$, where $(Q'_x, Q'_y, Q'_z)$ corresponds to the new coordinates of the mesh. These coordinates are subsequently utilized as input in $t + 1$ to predict the output parameters of the mesh for the next timestep, *i.e.*, for $t + 2$. Hence, the displaced coordinates are considered as Recurrent Parameters (RPs). In the example case presented in Fig. 4, the dimensions of the node branch output tensor would be $8×8×3×3$, where $8×8×3$ are the dimensions of the output tensor mesh and $3$ is the number of output features. As regards the element branch, the output tensor can be defined as $Y_{t+1}^e = (\Sigma, E)$, where $\Sigma$ corresponds to the effective stress and $E$ to the effective strain. Moreover, the element branch output tensor would be of shape $7×7×2×2$, where $7×7×2$ are the dimensions of the element tensor and 2 denotes the number of output parameters. Since $Y_{t+1}^e$ is not used as input for the next timestep predictions, the effective stress and strain are considered as Non-Recurrent Parameters (NRPs).

### 2.1.2    Feature Extraction Module

The FExM utilizes $r$ ConvLSTM layers, each responsible for extracting important spatiotemporal features of the FEA model that are identified by its kernels. Contrary to traditional CNN-based methods, the dimensionality of the input tensor is not reduced after each layer, thus the extracted feature map of each layer has the same dimensions as the input tensor with different number of features.

A ConvLSTM layer consists of three gates, *i.e.*, the input gate $\boldsymbol{i}_t$, the forget gate $\boldsymbol{f}_t$, and the output gate $\boldsymbol{o}_t$ (Fig. 5). These three gates are used to predict the cell state $\boldsymbol{C}_t$ and the hidden state $\boldsymbol{H}_t$ for the current timestep $t$. The input gate $\boldsymbol{i}_t$ is tasked to incorporate useful information from the input of the current timestep and the forget gate $\boldsymbol{f}_t$ is tasked to determine which piece of information should be retained from the memory cell status of the previous timestep. The output gate $\boldsymbol{o}_t$ determines whether the memory cell status of the current timestep will be propagated to the hidden state.

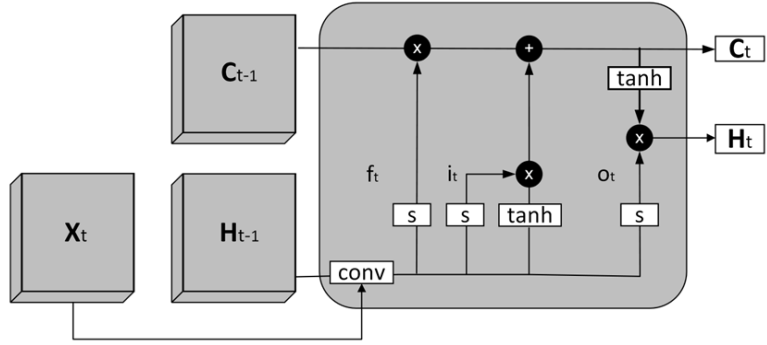

*Figure 5 Core architecture of a ConvLSTM layer.*

The cell of a ConvLSTM layer can be described by the following system of equations:

$$g(\cdot\,;\boldsymbol{W},\boldsymbol{b}) = \begin{cases} \boldsymbol{i}_t = s(\boldsymbol{W}_{xi} * \boldsymbol{X}_t + \boldsymbol{W}_{hi} * \boldsymbol{H}_{t-1} + \boldsymbol{W}_{ci} \circ \boldsymbol{C}_{t-1} + \boldsymbol{b}_i) \\ \boldsymbol{f}_t = s(\boldsymbol{W}_{xf} * \boldsymbol{X}_t + \boldsymbol{W}_{hf} * \boldsymbol{H}_{t-1} + \boldsymbol{W}_{cf} \circ \boldsymbol{C}_{t-1} + \boldsymbol{b}_f) \\ \boldsymbol{C}_t = \boldsymbol{f}_t \circ \boldsymbol{C}_{t-1} + \boldsymbol{i}_t \circ \tanh(\boldsymbol{W}_{xc} * \boldsymbol{X}_t + \boldsymbol{W}_{hc} * \boldsymbol{H}_{t-1} + \boldsymbol{b}_c) \\ \boldsymbol{o}_t = s(\boldsymbol{W}_{xo} * \boldsymbol{X}_t + \boldsymbol{W}_{ho} * \boldsymbol{H}_{t-1} + \boldsymbol{W}_{co} \circ \boldsymbol{C}_t + \boldsymbol{b}_o) \\ \boldsymbol{H}_t = \boldsymbol{o}_t \circ \tanh(\boldsymbol{C}_t) \end{cases} \quad (1)$$

where '$*$' denotes the convolution operation, '$\circ$' is the operator of the Hadamard product, $s(\cdot)$ denotes the sigmoid function, $\boldsymbol{W} = (\boldsymbol{W}_{xi}, \boldsymbol{W}_{hi}, \boldsymbol{W}_{xf}, \boldsymbol{W}_{hf}, \boldsymbol{W}_{xo}, \boldsymbol{W}_{ho}, \boldsymbol{W}_{xc}, \boldsymbol{W}_{hc})$ denotes the weight parametrization of $\boldsymbol{g}$, and $\boldsymbol{b}$ denotes its biases $(\boldsymbol{b}_i, \boldsymbol{b}_f, \boldsymbol{b}_o, \boldsymbol{b}_c)$. The tensors $(\boldsymbol{W}_{xi}, \boldsymbol{W}_{hi}, \boldsymbol{W}_{ci}, \boldsymbol{b}_i)$, $(\boldsymbol{W}_{xf}, \boldsymbol{W}_{hf}, \boldsymbol{W}_{cf}, \boldsymbol{b}_f)$, $(\boldsymbol{W}_{xo}, \boldsymbol{W}_{ho}, \boldsymbol{W}_{co}, \boldsymbol{b}_o)$, and $(\boldsymbol{W}_{xc}, \boldsymbol{W}_{hc}, \boldsymbol{b}_c)$ are the weight and bias tensors for the $\boldsymbol{i}_t$, $\boldsymbol{f}_t$, and $\boldsymbol{o}_t$ gates and the cell state $\boldsymbol{C}_t$, respectively. Moreover, $\boldsymbol{X}_t$ is the input tensor for the current timestep, $\boldsymbol{H}_t$ and $\boldsymbol{H}_{t-1}$ are the hidden states of the current and previous timesteps, respectively, and $\boldsymbol{C}_t$ and $\boldsymbol{C}_{t-1}$ are the memory cell statuses of the current and previous timesteps, respectively.

Each ConvLSTM layer $\boldsymbol{l}^j$ of the FExM, utilizes the predicted hidden state $\boldsymbol{H}_t^{j-1}$ of the previous layer. Based on Eq. (1), a layer $\boldsymbol{l}^j$ can be formally expressed as:

$$\boldsymbol{l}^j = g(\boldsymbol{H}_t^{j-1}, \boldsymbol{H}_{t-1}^j, \boldsymbol{C}_{t-1}^j; \boldsymbol{W}^j, \boldsymbol{b}^j) \; for \; j = 2, \ldots, r \quad (2)$$

where $g(\cdot; \boldsymbol{W}^j, \boldsymbol{b}^j)$ is Eq. (1) expressed for layer $\boldsymbol{l}^j$ and $r$ denotes the total number of ConvLSTM layers. In addition, $\boldsymbol{H}_{t-1}^j$ and $\boldsymbol{C}_{t-1}^j$ denote the hidden and cell states, respectively, for layer $\boldsymbol{l}^j$ and timestep $t$-1. For $j = 1$, Eq. (2) becomes:

$$\boldsymbol{l}^1 = g(\boldsymbol{X}_t, \boldsymbol{H}_{t-1}^1, \boldsymbol{C}_{t-1}^1; \boldsymbol{W}^1, \boldsymbol{b}^1) \quad (3)$$

Let $\boldsymbol{H}_t^r$ be the hidden state of timestep $t$ as estimated by the last ConvLSTM layer $\boldsymbol{l}^r$ of the FExM. The hidden state $\boldsymbol{H}_t^r$ is then propagated through each branch of the PM to predict the node- and element-related output parameters. This procedure is conducted sequentially for each simulation timestep, starting from the initial condition at $t_o = 0$ until the end of the simulation for a total number of $T$ timesteps.

### 2.1.3    Prediction Module

PM utilizes two CNN branches in a parallel topology. Each CNN branch comprises one or more CNN layers; in this study, one CNN layer is utilized for each branch. One branch is used to map the extracted feature maps to node tensors and the other is used to map the extracted feature maps to output tensors representing the elements of the FEA model. Furthermore, the CNN layers of each PM branch utilize the same kernel dimensions as the ConvLSTM layers of the FExM. Each branch is dedicated to predicting node- and element-related parameters. The distinction between the branches dedicated to node- and element-related predictions is mandatory, since they refer to outputs that are represented by tensors of different sizes. Therefore, for a 3D FEA model with hexahedral elements, the CNN layer of the node branch would utilize a $1\times1\times1\times K_n$ kernel, whereas that of the element branch would utilize a $2\times2\times2\times K_e$ kernel, with $K_n$ and $K_e$ denoting the number of output features for each branch, respectively.

The activation function of the CNN layers of each branch is determined by the data normalization process. For instance, if the input data are normalized within the interval of [-1, 1], then the hyperbolic tangent (*tanh*) can be considered as an appropriate activation function for each output layer of each branch. This is a requirement for DeepFEA, since the output of the NEP network at the timestep $t$ is used as input for the next timestep $t$+1.

## 2.2    Training Phase

The NEP network is trained using data generated from conventional FEA simulations, *i.e.*, using commercial or in-house software to simulate the behavior of a structure under different conditions, *e.g.*, applying forces with different magnitudes and angles on a structure of a given material. The training dataset contains representative FEA simulations of the examined model that will enable the NEP network to learn the examined behavior without being trained on the whole spectrum of possible FEA simulations. Subsequently, the trained NEP network can be used to predict the solutions for new FEA simulations of the examined FEA model with varying conditions, such as external load applied at different nodes of the mesh. Furthermore, the training phase relies on a NELO algorithm to address the error accumulation problem that occurs through the traditional training approach of deep learning models and a novel additive loss function to account for both node- and element-related errors produced by the NEP network.

### 2.2.1    Node-Element Loss Optimization Algorithm

The NELO algorithm is inspired by the SSM, which is an NLP technique used for the training of sequence-to-sequence models such as RNNs. These models generate an output text sequence (e.g., a translation or a text prediction) based on an input sequence (e.g., a source sentence). During training, the model uses the ground truth tokens (words or characters) at each step to predict the next token. However, during inference (generation of output sequences), the model uses its own predictions as inputs, which may lead to errors accumulating over time, similar to surrogate models of transient FEA simulations. Therefore, the NELO algorithm adopts a SSM optimization approach adapted to the domain of transient FEA simulations and tackles the error accumulation problem by gradually transitioning from using ground truth data to using model predictions as inputs during training. The NELO algorithm can be described as follows:

---

**Algorithm 1** NELO algorithm

---

1:     ***Set** $P_s \longleftarrow 1$, $k \longleftarrow$ number of epochs to decrease $P_s$, $S \longleftarrow$ total epochs, $T \longleftarrow$ total timesteps, $\kappa \longleftarrow 0$*

2:     ***for** $j = 0$ to S epochs*

3:       *Procedure Train model*

        ***for** $b = 0$ **to** B total batches*

4:           ***for** $t = 0$ **to** T total timesteps*

5:             *$P_r \longleftarrow$ random [0,1)*

6:             ***if** $P_r > P_s$ **and** $t > 0$*

7:               ***Then***

8:                 *Replace ground truth input with predicted output*

9:             *Predict output for next timestep*

10:           ***End***

11:         *Procedure Backpropagation, Optimizer step*

      ***End***

12:       ***if** $j$ % $k$ == 0*

        ***Then***

13:           *$\kappa \longleftarrow \kappa + 1$*

14:           *$P_s \longleftarrow$ Decrease by preferred scheme*

15:       ***if** $j < \beta_p$*

        ***Then***

16:           *$P_s$ == 0*

17:    ***End***

---

where $\beta_p \in (0, 1)$ is a constant and $P_s$ is the probability of using the ground truth data as input for the next timestep, which is defined as:

$$P_s = \gamma^\kappa \tag{8}$$

where $\gamma \in (0, 1)$ is a constant and $\kappa \in \mathbb{Z}^+$ can be described as an incremental factor, *i.e.*, an integer that is incremented by 1 every $k$ epoch. The $\gamma$ factor along with the incrementation rate $\kappa$ are adjusted according to the nature of the problem.

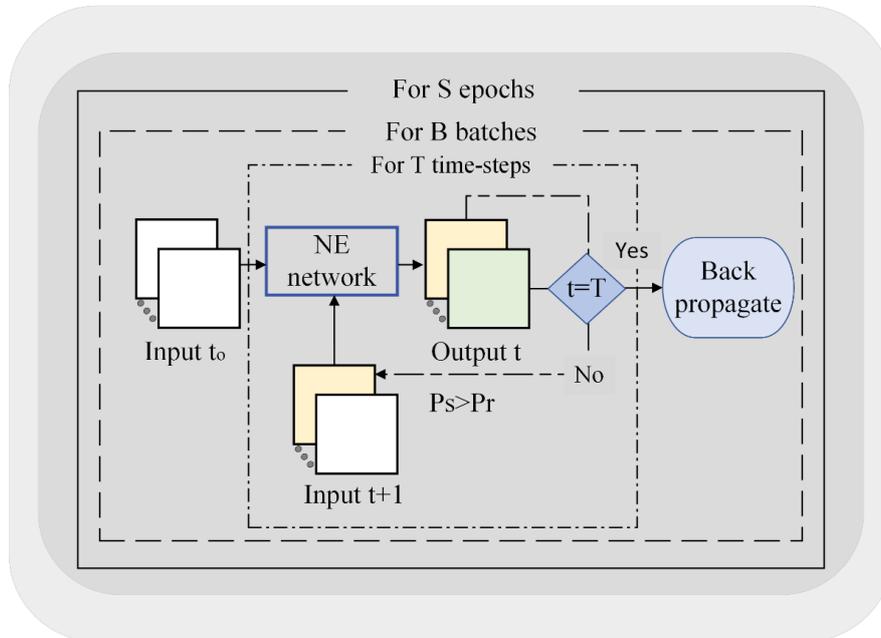

*Figure 6 Overview of the training phase.*

The training process is illustrated in Fig. 6, where, for each batch of FEA simulations with T timesteps, the NEP network is tasked to recurrently predict the output of each timestep $t$, starting from the initial timestep $t_o$. The predicted RPs (yellow tensors in Fig. 6) from timestep $t$ are used as input in timestep $t$+1 with a probability of 1-$P_s$ ($P_s$ > $P_r$, where $P_r$ is a random number $\in$ [0,1)). Based on Algorithm 1, at the beginning of training, the NEP network is forced to rely heavily on the ground truth. As training progresses, the NEP network starts using its own predictions more frequently. This gradual transition enables DeepFEA to learn to predict the solutions of the entire simulation more robustly and minimize errors.

### 2.2.2 Node-Element Loss Function

Considering the nature of the PM in the NEP network, the error of both the node and element branch need to be incorporated in the same loss function. Hence, a novel additive loss function, hereinafter called Node-Element (NE) loss, is devised. The formulation of the NE loss is inspired by the Mean Squared Error (MSE) metric; thus, it places greater emphasis on larger errors, accommodating discrepancies across the entire value spectrum of the output parameters and the squaring effect balances the impact of outliers, proving beneficial for handling noisy or irregularly distributed FEA simulation data [6]. The NE loss function employed in this study can be described as follows:

$$\mathcal{L}_{total} = \zeta_n \cdot \left( \frac{\sum_{n=1}^{N} \sum_{t=1}^{T} (y_{nt} - \hat{y}_{nt})^2}{T \cdot N} \right) + \zeta_e \cdot \left( \frac{\sum_{e=1}^{E} \sum_{t=1}^{T} (y_{et} - \hat{y}_{et})^2}{T \cdot E} \right) \qquad (9)$$

where $\zeta_n$ and $\zeta_e$ are the scaling factors for the node- and element-related output, respectively, $N$ is the total number of nodes, $E$ is the total number of elements, $T$ is the total number of timesteps, $\hat{y}_{nt}$ and $y_{nt}$ correspond to the predicted and ground truth output related to nodes, and $\hat{y}_{et}$ and $y_{et}$ correspond to the predicted and ground truth output related to elements.

By incorporating the predicted output tensor of both branches into an additive loss function, the network can concurrently learn to predict the output parameters for both nodes and elements without being limited by model-specific properties.

# 3  Experiments and Results

## 3.1  Dataset

For the purposes of this study, three different datasets of FEA simulations that predict the deformation of an object under different conditions were considered. In detail, two 2D model datasets utilizing material models with linear elastic (LEM) and hyperelastic (HM) characteristics, as well as one 3D LEM dataset were developed. The objects in each dataset were subjected to an external load applied at different outer nodes of the mesh under different angles and magnitudes over a period of 1 s. The software utilized for the generation of the datasets was ANSYS LS-DYNA. The outputs of all simulations were recorded every 5 ms, resulting in datasets comprising 200 datapoints, *i.e.*, 200 timesteps. The generated meshes contained 9×9 nodes and 64 elements for the 2D cases and 9×9×3 nodes and 128 elements for the 3D case. The force was randomly applied on one external node of each mesh and was linearly increased until the end of the simulation. The force was applied at four different angles {0°, 45°, 90°, 135°} and three different maximum force magnitudes were considered {5×10$^5$, 10$^6$, 2×10$^6$} N. The non-constrained nodes of the 2D and 3D meshes had 4- and 6-Degrees-Of-Freedom (DOFs), respectively. The bottom nodes of the meshes were set as constrained. The LEM and HM had a mass density of 1200 kg/m$^3$ each. The Young's modulus and Poison ratio of LEM were 5×10$^6$ Pa and 0.495, respectively. Regarding HM, the Ogden model was adopted with a Poisson's ratio, first shear modulus, and first exponent parameters of 0.495, 5.978×10$^4$ Pa, and 12.97, respectively. At each timestep, FEA was tasked to predict the deformation, *i.e.*, node displacement, and the effective stress and strain of the objects. This resulted in generating three different datasets containing 450 2D LEM, 450 2D HM, and 2,256 3D LEM simulation cases.

For the 2D dataset generation, all the combinations of angles and magnitudes were considered for all the external non-constrained nodes of the mesh, involving both tensile and compressive loading. As regards the 2D LEM dataset, the effective stress and strain values resided in the intervals [$1.9 \times 10^{-22}$, $2.9 \times 10^{6}$] Pa and [$3.07 \times 10^{-29}$, $5.6 \times 10^{-1}$], respectively. The displacement values ranged within [-0.046, 0.072] m for the x-axis and [$-5.2 \times 10^{-2}$, $2.3 \times 10^{-2}$] for the y-axis. As regards the 2D HM dataset, the effective stress and strain values resided in the intervals [$5.3 \times 10^{-6}$, $2.9 \times 10^{6}$] Pa and [$3.1 \times 10^{-20}$, $4.3 \times 10^{-1}$], respectively. The displacement values ranged within [$-9.3 \times 10^{-2}$, $9.3 \times 10^{-2}$] m for the x-axis and [$-5.8 \times 10^{-2}$, $5.8 \times 10^{-2}$] m for the y-axis. In this case of the 3D LEM dataset, the additional dimension (2 more DOFs) provided a larger dataset with compression forces applied to non-constrained nodes on the surface of the mesh, under the same angles and force magnitudes considered in the 2D dataset. The effective stress and strain values resided in the intervals [$4.86 \times 10^{-13}$, $5.38 \times 10^{6}$] Pa and [$9.7 \times 10^{-20}$, $1.7 \times 10^{-1}$], respectively. The displacement values ranged within [-0.13, 0.13] m for the x-axis, [$-1.3 \times 10^{-1}$, $9 \times 10^{-2}$] m for the y-axis, and [$-1.7 \times 10^{-2}$, $1.9 \times 10^{-1}$] m for the z-axis.

In this study, the effective stress and strain are calculated based on the von Mises equation that can be described as:

$$\sigma = \sqrt{\frac{(\sigma_{xx} - \sigma_{yy})^2 + (\sigma_{yy} - \sigma_{zz})^2 + (\sigma_{zz} - \sigma_{xx})^2 + 6(\sigma_{xy}{}^2 + \sigma_{yz}{}^2 + \sigma_{zx}{}^2)}{2}} \qquad (6)$$

$$\varepsilon = \frac{2}{3} \sqrt{\frac{(\varepsilon_{xx} - \varepsilon_{yy})^2 + (\varepsilon_{yy} - \varepsilon_{zz})^2 + (\varepsilon_{zz} - \varepsilon_{xx})^2 + 6(\varepsilon_{xy}{}^2 + \varepsilon_{yz}{}^2 + \varepsilon_{zx}{}^2)}{2}} \qquad (7)$$

where $\sigma$ denotes the effective stress, ($\sigma_{xx}, \sigma_{yy}, \sigma_{zz}, \sigma_{xy}, \sigma_{yz}, \sigma_{zx}$) are the stress tensor components, $\varepsilon$ denotes the effective strain, and ($\varepsilon_{xx}, \varepsilon_{yy}, \varepsilon_{zz}, \varepsilon_{xy}, \varepsilon_{yz}, \varepsilon_{zx}$) are the strain tensor components for each element of the FE mesh. In the 2D scenario, the stress components ($\sigma_{zz}, \sigma_{yz}, \sigma_{zx}$) and strain components ($\varepsilon_{zz}, \varepsilon_{yz}, \varepsilon_{zx}$) were zero.

### 3.2 Evaluation metrics

To quantify the effectiveness of DeepFEA, the $R^2$ metric, known as coefficient of determination, has been selected, since it is widely used for the evaluation of regression models [43]. $R^2$ is a statistical measure that indicates the goodness of fit of a regression model, *i.e.*, how well the model's predictions align with the ground truth values. $R^2$ is usually confined in the [0,1] interval, where 1 indicates that the predictions of a model perfectly fit the ground truth data, whereas 0 indicates a baseline model that always predicts the mean of the ground truth data. $R^2$ can take negative values in cases where the model is worse than the baseline model. $R^2$ is defined as:

$$R^2 = 1 - \frac{\sum_j (y_j - \hat{y}_j)^2}{\sum_j (y_j - \overline{y}_j)^2} \qquad (10)$$

where $y_j$ is the $j^{th}$ ground truth value, $\hat{y}_j$ is the $j^{th}$ predicted value, and $\overline{y}$ is the mean of all ground truth values.

The analysis of output parameters, such as stresses and strains, presents a challenge due to their varying order of magnitude, *e.g.*, stress values may span from Pa to GPa. Hence, the normalized Mean Absolute Error (NMAE) and Root Mean Squared Error (NRMSE) have been incorporated into our evaluation study. NMAE and NRSMSE have been reported to be more robust compared to MAE and RMSE when assessing the performance of predictors in similar scenarios [6].

$$NMAE_p = \frac{1}{A} \sum_{j=1}^{A} \frac{\sum_{\mu=1}^{\Gamma} |y_{\mu j} - \hat{y}_{\mu j}|}{\Gamma \cdot (max\{Y_j^p\} - min\{Y_j^p\})} \times 100\% \qquad (11)$$

$$NRMSE_p = \frac{1}{A} \sum_{j=1}^{A} \frac{\sqrt{\sum_{\mu=1}^{\Gamma} (y_{\mu j} - \hat{y}_{\mu j})^2}}{\sqrt{\Gamma} \cdot (max\{Y_j^p\} - min\{Y_j^p\})} \times 100\% \qquad (12)$$

where $y_{\mu j}$ is the ground truth output of the $\mu^{th}$ node or element for the $j^{th}$ simulation, $\hat{y}_{\mu j}$ is the predicted value of the $\mu^{th}$ node or element for the $j^{th}$ simulation, $A$ is the total number of simulations, $\Gamma$ is the total number of ground truth output values for simulation $j$, $max\{Y_j^p\}$ and $min\{Y_j^p\}$ are the maximum and minimum values, respectively, of the set $Y$ containing the output values of the $j^{th}$ simulation for the parameter $p$.

### 3.3 Experimental Setup

The performance of the method was evaluated in the context of predicting the effective stresses and strains and the displacement ($d_x$, $d_y$, $d_z$) in each Cartesian axis (x, y, and z (for 3D)). The displacements predicted for each axis were also utilized to calculate the resultant displacement ($R_d$) of the nodes. An ablation study was conducted based on the 2D LEM dataset to determine the best combination of layers and channels using a 3×3 kernel size for the convolutional layers. The best architecture was selected based on the quantitative evaluation metrics. Subsequently, the selected architecture was used to demonstrate the capabilities of DeepFEA across a variety of FEA simulation scenarios. More specifically, DeepFEA was trained on the 2D LEM, 3D LEM, and 2D HM datasets, and the results were assessed both quantitatively and qualitatively.

To evaluate the performance of DeepFEA, the datasets were divided into training and testing subsets with proportions of 80% and 20%, respectively. The networks for the 2D and 3D FEA simulation predictions were trained with a batch size of 32 and 16, respectively. The optimizer used in the experiments was the Adam optimizer with a variable learning rate associated with the value of the NELO factor $P_s$. The scaling factors $\zeta_n$ and $\zeta_e$ of the loss function were set to $10^4$. Moreover, the $\gamma$ factor of the NELO algorithm was set to 0.7, $\beta_p$ was set to 0.01, and the $P_s$ was reduced every 40 epochs. The model was implemented using the PyTorch v1.12 framework [44] and it was trained with a GeForce RTX 3090 24GB.

### 3.4 Ablation study

An ablation study was conducted to identify the best architecture through extensive experimentation. DeepFEA was trained on a subset of the 2D LEM dataset with different numbers of layers of varying channel sizes, whereas the kernel size was kept constant to 3×3. As it can be seen in Table 1, the overall best model comprised three ConvLSTM layers with 64, 128, and 256 channels. It can be observed that DeepFEA can consistently predict the effective stress and strain of the FEA simulations with great accuracy across all timesteps, with the best model achieving an $R^2$ of ~0.99. It is worth noting that, according to the results presented in Table 1, the node displacement prediction is a more complex task and thus the utilization of a network with higher complexity benefits the overall performance. Nevertheless, based on the same results, it can be observed that an architecture with more layers is not necessarily beneficial for the performance of the method, *i.e.*, the three layer variant outperforms the four layer ones.

*Table 1 Quantitative results of the ablation study for models with different ConvLSTM layers trained on a subset of the 2D LEM dataset. Best performance is indicated in bold.*

| Architectures | | 1 layer 64 (3×3) | 1 layer 128 (3×3) | 1 layer 256 (3×3) | 2 layers 64 (3×3)-64 (3×3) | 2 layers 64 (3×3)-128 (3×3) | 2 layers 128 (3×3)-128 (3×3) | 2 layers 128 (3×3)-256 (3×3) | 3 layers 64 (3×3)-128 (3×3)-256 (3×3) | 4 layers 64 (3×3)-64 (3×3)-128 (3×3)-256 (3×3) | 4 layers 64 (3×3)-128 (3×3)-128 (3×3)-256 (3×3) |
|---|---|---|---|---|---|---|---|---|---|---|---|
| **$R^2$ ↑** | $\sigma$ | 0.964 | 0.990 | 0.990 | 0.991 | 0.982 | 0.988 | **0.994** | 0.993 | 0.991 | 0.993 |
| | $\varepsilon$ | 0.962 | 0.989 | 0.989 | 0.990 | 0.981 | 0.988 | **0.994** | 0.993 | 0.991 | 0.993 |
| | $d_s$ | 0.445 | 0.858 | 0.918 | 0.927 | 0.833 | 0.934 | 0.949 | 0.963 | 0.954 | **0.965** |
| | $d_{\theta}$ | 0.694 | 0.893 | 0.939 | 0.903 | 0.898 | 0.954 | 0.957 | **0.975** | 0.969 | 0.974 |
| | $R_d$ | 0.774 | 0.935 | 0.967 | 0.961 | 0.931 | 0.971 | 0.980 | **0.984** | 0.976 | 0.980 |
| **NMAE (%) ↓** | $\sigma$ | 1.509 | 0.802 | 0.755 | 0.698 | 1.035 | 0.778 | 0.559 | **0.511** | 0.544 | 0.535 |
| | $\varepsilon$ | 0.916 | 0.858 | 0.806 | 0.740 | 1.090 | 0.833 | 0.601 | **0.544** | 0.610 | 0.570 |
| | $d_s$ | 5.324 | 2.866 | 2.264 | 2.225 | 3.289 | 2.078 | 1.778 | 1.513 | 1.764 | **1.481** |
| | $d_{\theta}$ | 3.645 | 2.124 | 1.628 | 2.206 | 2.106 | 1.437 | 1.381 | **0.986** | 1.088 | 1.005 |
| | $R_d$ | 3.618 | 2.009 | 1.554 | 1.690 | 2.199 | 1.455 | 1.201 | 1.034 | 1.679 | **1.019** |
| **NRMSE (%) ↓** | $\sigma$ | 2.268 | 1.175 | 1.169 | 1.092 | 1.573 | 1.226 | 0.878 | **0.853** | 0.964 | 0.904 |
| | $\varepsilon$ | 2.37 | 1.256 | 1.225 | 1.136 | 1.644 | 1.297 | 0.914 | **0.887** | 0.989 | 0.955 |
| | $d_s$ | 8.109 | 4.082 | 3.224 | 3.049 | 4.500 | 2.915 | 2.449 | 2.194 | 2.397 | **2.135** |
| | $d_{\theta}$ | 5.049 | 2.917 | 2.280 | 2.836 | 2.944 | 2.016 | 1.858 | **1.451** | 1.606 | 1.491 |
| | $R_d$ | 5.434 | 2.880 | 2.195 | 2.272 | 3.062 | 2.036 | 1.656 | 1.506 | 1.679 | **1.473** |

Note: ↑ indicates better performance for larger values and ↓ for smaller ones.

To further evaluate the proposed method, the ConvLSTM layers of DeepFEA were replaced with simple CNN layers, Bi-LSTM layers, and a variation of the GNN-GRU. These networks failed to provide comparable results for every parameter and metric. More specifically, the CNN- and GNN-GRU-based networks achieved worse $R^2$ values ($R^2$ << -1) than the baseline ($R^2$ = 0) and yielded counterproductive results according to the NMAE and NRMSE metrics (NMAE and NRMSE >>100%) for all the output parameters. The effective strain and stress predicted by the Bi-LSTM-based model ($R^2$ = ~0.3, NMAE = ~6%, and NRMSE = ~9%) were slightly better than those predicted by the other networks, but significantly worse than those predicted by DeepFEA ($R^2$ = ~0.99, NMAE = ~0.6%, and NRMSE = ~0.9%). As regards the predicted displacement, the performance of Bi-LSTM was similar to that of the other two ANNs ($R^2$ << -1, NMAE & NRMSE >>100%). Since these ANNs performed poorly, the following subsections will focus on the results produced by DeepFEA for each dataset.

Deep learning models require a significant amount of training samples and may perform poorly in terms of generalizability. For transient FEA simulations, it is essential that once the deep learning model has been trained, it can accurately predict the solution of new simulation cases (not included in the training set) for the same FEA model. To determine the optimal trade-off between accuracy and dataset generation time, DeepFEA was trained with different ratios of training and testing samples of the 2D LEM dataset. The results presented in Table 2 and Fig. 7 indicate that DeepFEA is capable of accurately predicting the effective stress and strain even when trained on 40% of the dataset. It can also be observed that the performance of DeepFEA regarding the predicted displacements is negatively impacted when the NEP network is trained on less than 60% of the dataset.

*Table 2 Quantitative results of the ablation study for training subsets with different percentages (2D LEM dataset). Best performance is indicated in bold.*

| Metrics | | % of Dataset | | | |
|---|---|---|---|---|---|
| | | 20 | 40 | 60 | 80 |
| $R^2 \uparrow$ | $\sigma$ | 0.873 | 0.945 | 0.977 | **0.993** |
| | $\varepsilon$ | 0.878 | 0.944 | 0.976 | **0.993** |
| | $d_x$ | 0.614 | 0.692 | 0.890 | **0.963** |
| | $d_y$ | 0.569 | 0.792 | 0.926 | **0.975** |
| | $R_d$ | 0.765 | 0.874 | 0.945 | **0.984** |
| NMAE (%) $\downarrow$ | $\sigma$ | 2.122 | 1.341 | 0.882 | **0.511** |
| | $\varepsilon$ | 2.193 | 1.440 | 0.937 | **0.544** |
| | $d_x$ | 4.776 | 3.656 | 2.635 | **1.513** |
| | $d_y$ | 3.836 | 2.687 | 1.637 | **0.986** |
| | $R_d$ | 3.738 | 2.701 | 1.833 | **1.034** |
| NRMSE (%) $\downarrow$ | $\sigma$ | 3.943 | 2.385 | 1.518 | **0.853** |
| | $\varepsilon$ | 3.969 | 2.499 | 1.589 | **0.887** |
| | $d_x$ | 7.214 | 5.511 | 3.695 | **2.194** |
| | $d_y$ | 5.899 | 4.015 | 2.502 | **1.451** |
| | $R_d$ | 5.627 | 3.942 | 2.649 | **1.506** |

Note: ↑ indicates better performance for larger values and ↓ for smaller ones.

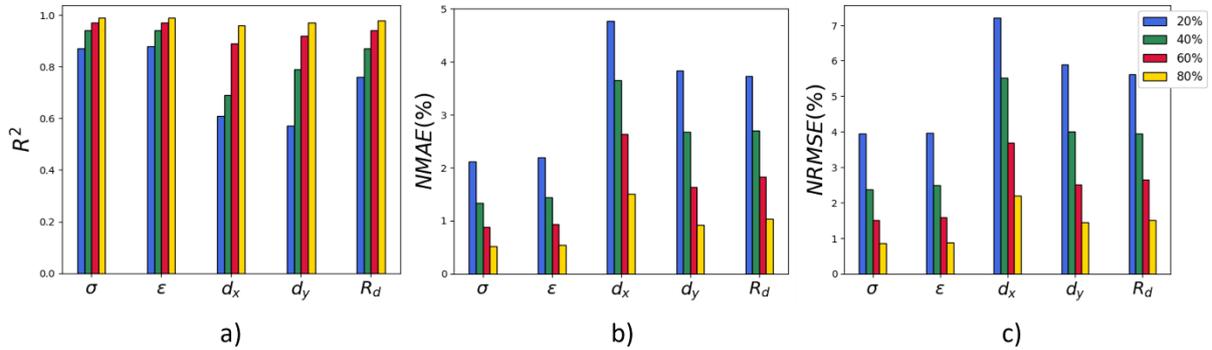

*Figure 7 Plots of the results presented in Table 2 for a) $R^2$, b) NMAE and c) NRMSE under training subsets of 20%, 40%, 60%, and 80%.*

## 3.5    2D Datasets (shell elements)

### 3.5.1    LEM Dataset

Based on the results of the ablation study, the model with the best architecture using 80% of the dataset was used for the quantitative and qualitative evaluation of the method. The results in Table 2 indicate that DeepFEA was able to predict all the parameters with great accuracy, exhibiting great consistency throughout the simulation ($R^2 > 0.95$). The values of $d_x$ were predicted with less accuracy compared to those of $d_y$, which can be attributed to the wider displacement range in the x-axis. Although the predictions for $R_d$ were similar to those for $d_y$, the normalized errors were higher due to the increased error in the x-axis.

To qualitatively evaluate the predictions of DeepFEA, three simulation cases were randomly selected from the dataset. Figure 8 presents representative snapshots (five timesteps) of the output parameters (deformation, strain, and stress evolution) predicted for each simulation case.

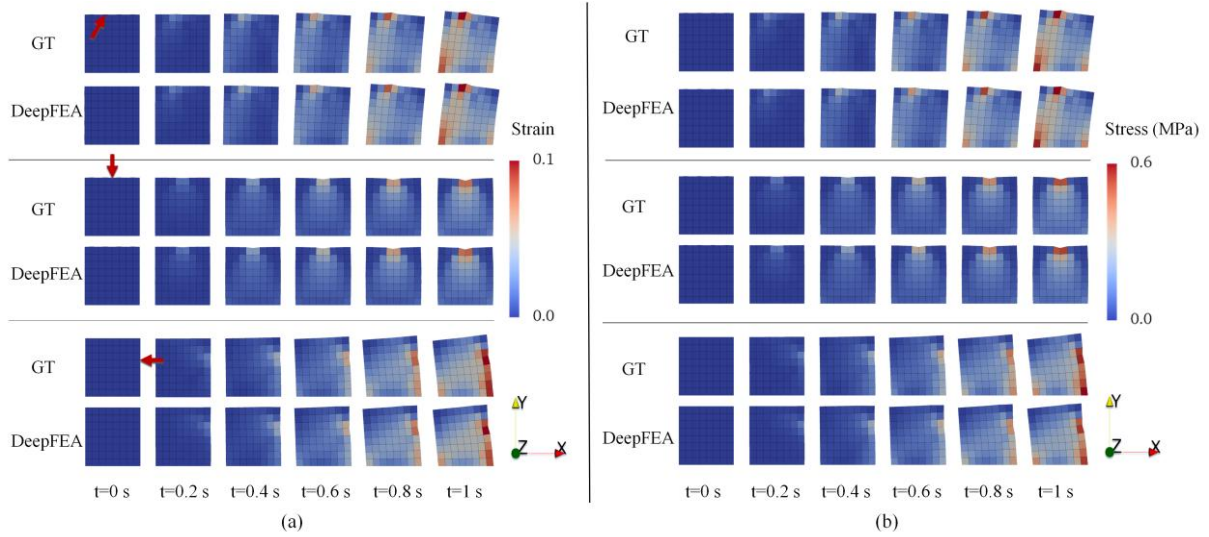

*Figure 8. Mesh deformation and (a) strain and (b) stress contours of three randomly-selected simulation cases (2D LEM dataset). The red arrows indicate the position and direction of the applied force. GT denotes ground truth.*

In addition, the evolution of the predicted average displacement (Fig. 9a), strain (Fig. 9b), and stress (Fig. 9c) of all nodes and elements for the entire simulation duration was plotted against the ground truth (dashed lines). Figure 9 depicts the average output of the third simulation case in Fig. 8 that was selected as a representative example. Overall, the results demonstrate the ability of DeepFEA to predict the solution of FEA simulations for different ranges of motion generated by forces applied at different external nodes of the 2D mesh.

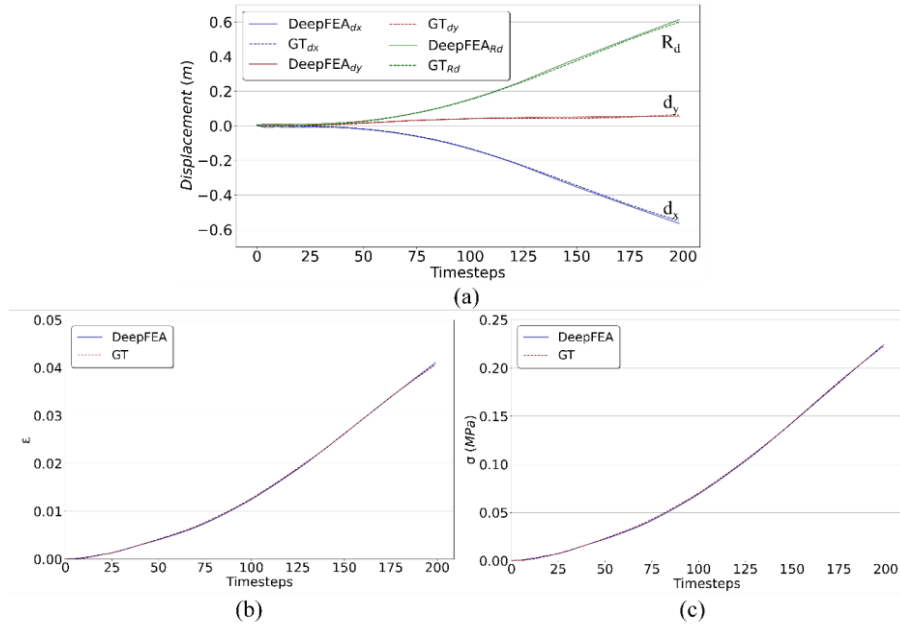

*Figure 9. Plots of (a) average displacement, (b) strain, and (c) stress vs. time for the third simulation case in Fig. 8 (2D LEM dataset).*

### 3.5.2 HM Dataset

DeepFEA was also tasked to predict the solutions of simulation cases from the 2D HM dataset. The quantitative results in Table 3 further validate the ability of DeepFEA to accurately predict the simulation outcome of FEA models with non-linear material properties. Notably, the predicted displacement was on par with the results in Table 2, with the only exception that the normalized errors for the displacement in the x- and y-axes were slightly higher. For this dataset, the normalized errors for the effective stress and strain differed from those in Table 2. More specifically, the task of predicting the effective strain in the HM dataset appeared to be more challenging than that in the LEM dataset. Despite that, the effective stress predicted by DeepFEA for the HM dataset achieved only a slightly lower $R^2$ and higher NRMSE compared to those for the LEM dataset (Table 2).

*Table 3 Quantitative results for DeepFEA trained on the 2D HM dataset. The metrics were obtained by comparison to the ground truth.*

| Metrics | DeepFEA | | | | |
|---|---|---|---|---|---|
| | Simulation Parameters | | | | |
| | $\sigma$ | $\varepsilon$ | $d_x$ | $d_y$ | $R_d$ |
| $R^2 \uparrow$ | 0.981 | 0.988 | 0.959 | 0.950 | 0.976 |
| NMAE (%) ↓ | 0.651 | 1.109 | 1.663 | 1.405 | 1.270 |
| NRMSE (%) ↓ | 1.278 | 1.732 | 2.587 | 2.323 | 1.982 |

Note: ↑ indicates better performance for larger values and ↓ for smaller ones.

### 3.6   3D LEM Dataset (solid elements)

To assess the capability of the proposed method in the 3D domain, DeepFEA was also evaluated on the 3D LEM dataset. The quantitative evaluation is summarized in Table 4. Despite the increased complexity of the problem due to the additional DOFs, DeepFEA was able to accurately predict $\sigma$ and $\varepsilon$, with $R^2 = \sim$0.99, NMAE = $\sim$0.49%, and NRMSE = $\sim$0.78%. DeepFEA inferred the resultant displacement of the nodes with high accuracy. Notably, the predicted $d_y$ was less accurate in terms of $R^2$ compared to the other displacement axes, while the normalized errors were comparable. In addition, the predicted $R_d$ demonstrated a closer alignment with the ground truth for all metrics. Thus, it can be inferred that, while the predicted displacement along a specific axis might deviate from the ground truth, DeepFEA can accurately capture the resultant deformation.

*Table 4 Quantitative results for DeepFEA trained on the 3D LEM dataset. The metrics were obtained by comparison to the ground truth.*

| Metrics | DeepFEA | | | | | |
|---|---|---|---|---|---|---|
| | Simulation Parameters | | | | | |
| | $\sigma$ | $\varepsilon$ | $d_x$ | $d_y$ | $d_z$ | $R_d$ |
| $R^2 \uparrow$ | 0.989 | 0.989 | 0.888 | 0.825 | 0.894 | 0.959 |
| NMAE (%) ↓ | 0.491 | 0.490 | 1.589 | 1.344 | 1.487 | 1.020 |
| NRMSE (%) ↓ | 0.784 | 0.784 | 2.227 | 1.971 | 2.101 | 1.406 |

Note: ↑ indicates better performance for larger values and ↓ for smaller ones.

Figure 10 illustrates representative snapshots (five timesteps) of the predicted output parameters (deformation, strain, and stress evolution) of three simulation cases that were randomly selected from the dataset. In the first case, a compression force was applied on an outer node of the mesh with magnitude of $10^6$ N, and angle of 0° relative to the z-axis. In the second case, a compression force was applied with a magnitude of $10^6$ N and angle of 135° with respect to the x- and y-axis. In the third case, a compression force was applied with a magnitude of $5 \times 10^5$ N and angle of 0° with respect to the x-axis.

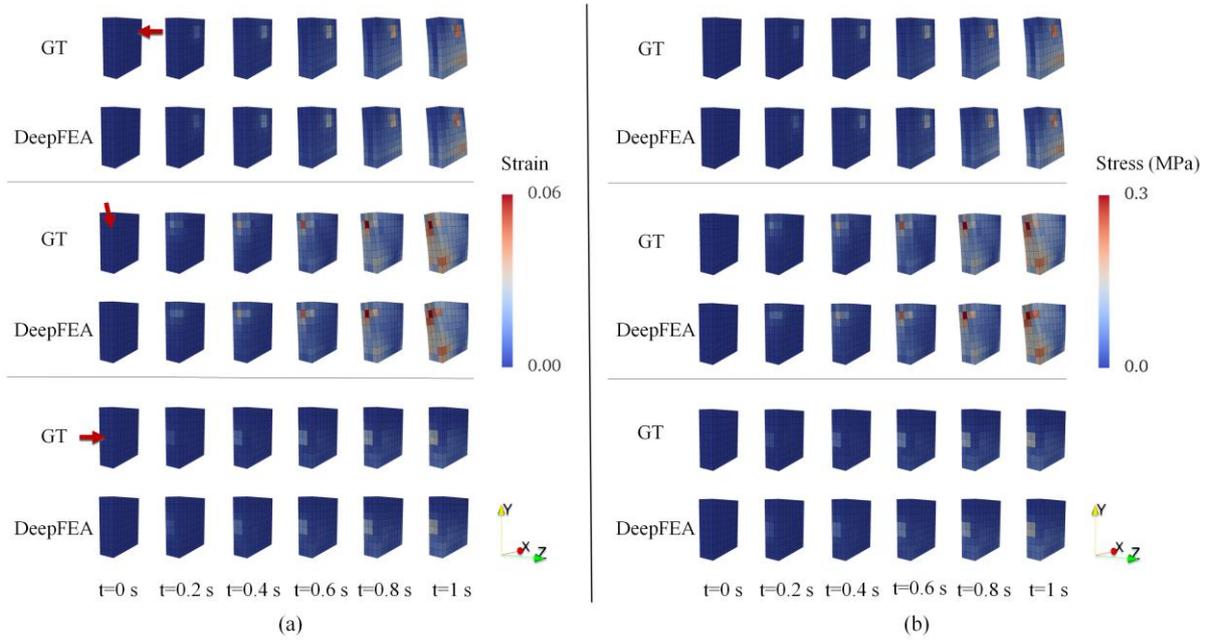

*Figure 10. Mesh deformation and (a) strain and (b) stress contours of three randomly-selected simulation cases (3D LEM dataset). The red arrows indicate the position and direction of the applied force. GT denotes ground truth..*

Furthermore, the second simulation case presented in Fig. 11 was selected as an example and the evolution of the average displacement (Fig. 11a), strain (Fig. 11b), and stress (Fig. 11c) of all nodes and elements for the entire simulation duration predicted by DeepFEA was plotted against the ground truth.

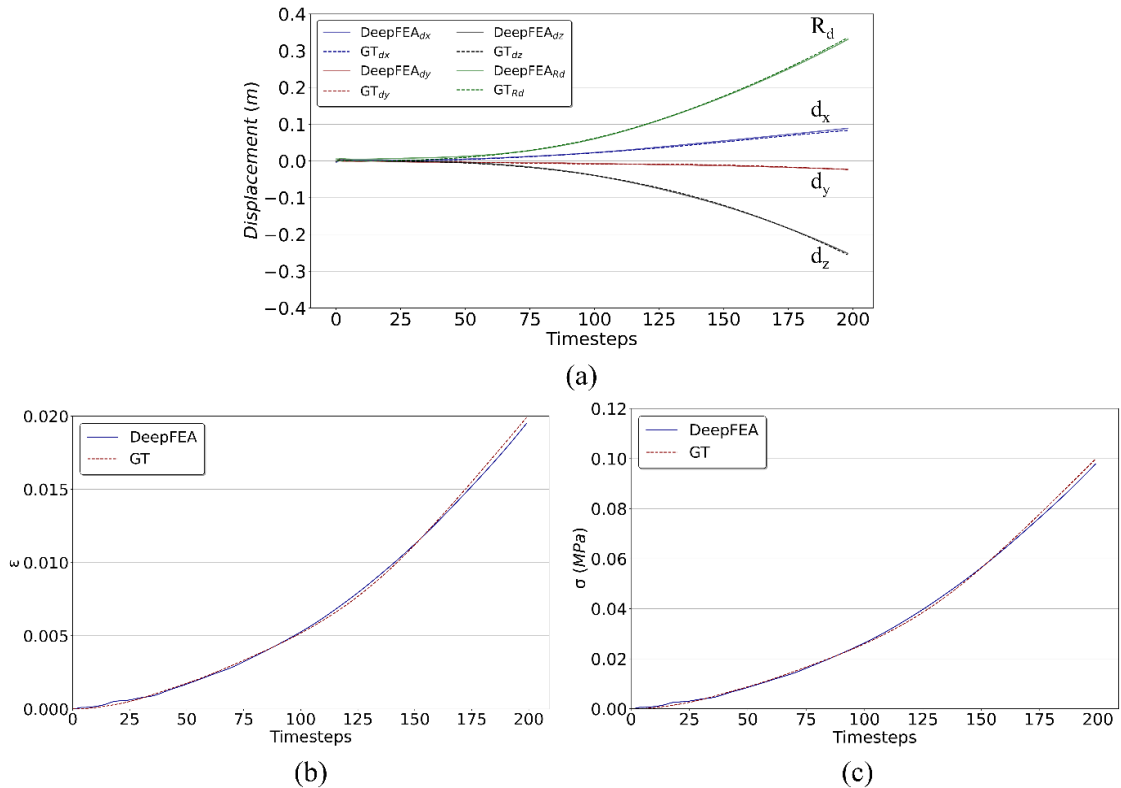

*Figure 11. Plots of (a) average displacement, (b) strain, and (c) stress vs. time for the second simulation case in Fig. 10 (3D LEM dataset).*

The effect of the constrained nodes on the y-axis displacement can be observed in Figs. 10 and 11a. The deformation along the y-axis is smaller than that in the x- and z-axes due to the limited range of motion. This could explain the discrepancy between the $R^2$ and normalized errors in Table 4 for the y-axis. Moreover, the minor error margins of the predicted $\sigma$ and $\varepsilon$ in Table 4 can be attributed to overestimations of the values of certain elements near the applied force position at the last timesteps of the simulation or due to marginal deviations throughout the entire simulation.

### 3.7 Inference Time

The results have demonstrated the performance of DeepFEA in terms of accuracy. Hence, a comparative analysis of the average CPU- and GPU-enabled inference times of DeepFEA and the CPU-enabled data generation time of FEA for each experimental scenario was conducted (Table 5). The resulted average inference times were obtained on a computer system with an Intel(R) Core(TM) i7-9750H CPU at 2.60GHz, a GeForce RTX 3090 24GB, and a 16 GB RAM. As regards the 2D datasets, DeepFEA could predict the solution of a FEA simulation in 0.20 s and 2.35 s on average using a GPU and CPU, respectively, whereas FEA needed 24 s. As regards the 3D LEM dataset, the inference time of DeepFEA for one simulation was 0.23 s (GPU) and 15.30 s (CPU) on average, whereas FEA required 40 s. This indicates that under the same conditions and the utilization of a GPU, the proposed method is up to two orders of magnitude faster than FEA for the 2D and 3D scenarios. Furthermore, the average inference time of DeepFEA was consistent for both the 2D and 3D domains when a GPU was employed, in contrast to the average solution time of FEA that exhibited a two-fold increase.

*Table 5 Average CPU and GPU enabled inference time (in seconds) of DeepFEA compared to CPU enabled generation time of FEA for one simulation.*

| Methods | Datasets | | | | | |
|---------|----------|----------|----------|----------|----------|----------|
| | 2D LEM | | 2D HM | | 3D LEM | |
| | *CPU (s)* | *GPU (s)* | *CPU (s)* | *GPU (s)* | *CPU (s)* | *GPU (s)* |
| *FEA* | 22 | *N/A* | 25 | *N/A* | 40 | *N/A* |
| *DeepFEA* | 2.35 | 0.20 | 2.35 | 0.20 | 15.30 | 0.23 |

## 4 Discussion

In this study, a novel deep learning surrogate model for transient FEA simulations has been proposed. FEA-related studies have focused on surrogate models for steady-state analysis [2, 7, 16, 19–22, 24]. Nevertheless, there have been only few early studies on surrogate models that are aimed for transient FEA simulations and have provided insight into the current progress and limitations [12, 30–33, 36, 39]. Limitations of these approaches include failure to account for spatiotemporal features, applicability limited to the 2D domain, and dependency on FEA for calculating part of the solution. To address these issues, DeepFEA was designed to predict the solutions of the entire transient FEA simulation given only the initial conditions, the external load, and the boundary conditions.

Mesh deformation is a crucial factor involved in structural mechanics [6, 12, 17, 19, 21, 39]. In transient FEA simulations, the deformation of the mesh depends on the predictions of the previous timesteps. Therefore, the core problem addressed by DeepFEA is the efficient prediction of RPs (here, node displacement). Related studies have proposed solutions for predicting such parameters, but are limited to the 2D domain and account for single-output predictions [36]. DeepFEA provides a framework that is capable of accurately predicting both RPs and NRPs of FEA simulations in the 2D and 3D domains. This is achieved by the implementation of a data-driven approach that incorporates the physical properties of the model in the input tensor (coordinates of the nodes, external load, and boundary conditions) and the utilization of the SSM training scheme. Thus, DeepFEA can be trained for FEA simulations with an arbitrary number of input and output parameters. In addition, DeepFEA is not limited to a specific problem, and, with proper adjustments, it can be applied to FEA models with various characteristics and properties.

The proposed approach of incorporating the coordinates as input features gives an additional merit to DeepFEA. As the structure of the mesh changes due to deformation, the connections between the nodes remain the same, whereas the input channels related to the coordinates of the nodes change to reflect the deformation of the mesh; this enables DeepFEA to solve the regression problem. Thus, this approach can be applied to unstructured meshes by mapping the nodes and elements of the mesh to appropriate tensors similarly to the examined cases of distorted FE meshes due to deformation. As regards the internal load distribution, related studies have either used the loads calculated by FEA as input [33, 35] or tried to approximate the effect of the external load on each node via elaborate approximation schemes [17]. On the contrary, DeepFEA relies only on the external load, which demonstrates that it is capable of understanding the complex physical interactions between nodes and encompasses the approximation of the internally applied load in its weights. Moreover, the comparison of different ANNs in Subsection 3.4 emphasizes the beneficial effect of the ConvLSTM layers in the FExM of DeepFEA. Although the examined networks (CNNs, LSTMs, and GNN-GRU) have been successfully employed in different FEA simulation domains (*e.g.*, structural analysis [12, 22, 32], FSI [33]), they are not able to simultaneously predict the deformation, stress, and strain of a mesh that is subjected to an external load over a period of time. This further highlights the need for building surrogate models that can generalize to different domains, such as structural mechanics, fluid dynamics, and FSI, without extensive modifications, *i.e.*, utilizing different ANN architectures for each FEA application. In this way, the evaluation of such models could be standardized, enabling a more efficient validation and comparison process.

DeepFEA was evaluated against three datasets, *i.e.*, 2D LEM, 2D HM, and 3D LEM. These datasets were selected for the performance evaluation of the proposed model in the 2D and 3D domains, as well as to assess the capability of DeepFEA to predict the solutions for models with non-linear material properties. The evaluation results indicate that DeepFEA is capable of accurately predicting the solutions for both the 2D and 3D LEM datasets. The quantitative results provided by DeepFEA (Tables 2 and 4) indicate its capacity to provide consistently accurate predictions during the entire simulation. This is also reflected in the qualitative results (Figs. 8-11). The NRMSE and NMAE for the displacement prediction of each node was estimated to be less than 3%, whereas its predictions regarding $\sigma$ and $\varepsilon$ were below 1%. According to the quantitative results, the 3D scenario is more demanding as regards the prediction of the mesh deformation. This can be attributed to the additional DOFs, *i.e.*, translation and rotation along the z-axis, which make the deformation characteristics of the mesh more complex. Nevertheless, the results obtained for the 3D case were comparable to those for the 2D one (Table 4). Furthermore, the ablation study revealed that DeepFEA is capable of accurately predicting the effective stress and strain of new FEA simulation cases using only 40% of the generated dataset to train the NEP network, while 60% is sufficient for the displacements. Depending on the examined problem and the available computational resources, DeepFEA can be trained on a smaller dataset of FEA simulation cases while maintaining sufficient accuracy. This could be beneficial in cases of FEA simulations that require significant computational time to be generated.

Apart from its capacity to be applied on both the 2D and 3D domains, DeepFEA can accurately predict the solutions of simulations involving hyperelastic models. Therefore, it can identify the complex characteristics of different models and differentiate between simulations involving linear or non-linear responses. It should be noted that in the context of the HM dataset, the prediction accuracy of DeepFEA was higher with respect to the stresses. On the other hand, the results for the LEM dataset exhibited similar performance both for $\varepsilon$ and $\sigma$. This may stem from the non-linear characteristics of the material model and the CNN layers in the element branch of DeepFEA, where the same kernels are used to predict the output for the effective stress and strain. During training, the loss function of the element branch generates the average error for both $\sigma$ and $\varepsilon$. In the case of HM, due to the non-linearity, $\varepsilon$ produces more complex outputs than $\sigma$, which may cause the weights of the kernels to converge on a solution that generates a larger error for $\varepsilon$. This averaging effect can also be observed in the LEM dataset, where $\sigma$ and $\varepsilon$ converged with the same error. Hence, in future studies, it would be beneficial to examine different loss functions, *e.g.*, incorporating different weights for each variable, that can mitigate this effect. As presented in Subsection 3.7, the inference time for both the 2D and 3D datasets was significantly lower than the FEA solution time. Therefore, DeepFEA can substantially reduce the computational time needed for studies that examine specific behaviors under different conditions.

# 5  Conclusions

This work presented a novel deep learning approach for predicting the solution of transient FEA simulations. DeepFEA is able to recurrently predict the output parameters for all timesteps in a simulation by utilizing a ConvLSTM-based network with two parallel convolutional branches and an NLP-inspired training scheme. To the best of our knowledge, this is the first time that the SSM training scheme has been used for predicting outputs of FEA simulations. The proposed method was evaluated in the domain of structural mechanics via 2D and 3D FEA models, as well as for two different material models (linear elastic and hyperelastic). The results indicated that DeepFEA can predict the evolution of the output parameters with great accuracy in all scenarios, being up to two orders of magnitude faster than FEA. The efficient inference time combined with the accurate predictive performance of DeepFEA in estimating solutions for FEA simulations across multiple timesteps underscores that, after training, it is able to predict the evolution of RPs and NRPs in a simulation without the assistance of FEA. DeepFEA can be trained to predict the solutions of several simulations involving models with varying boundary conditions and external loads, handle distorted meshes (caused by deformation), and adapt to different material models. Moreover, the ablation study showcased the potential of DeepFEA to accurately predict the solutions of FEA simulations when trained with fewer training samples. In addition, DeepFEA does not require the provision of pre-calculated or approximated internal loads. This achievement lays a robust foundation for future research in the field of transient FEA by utilizing similar approaches. Despite its advantages, DeepFEA can be trained to predict the solutions of transient FEA simulations for FEA models with the same mesh size and density. This limitation is due to the fixed dimensions of the weights in the ConvLSTM layers of the FExM. Thus, a different NEP network needs to be trained each time for FEA models with different mesh characteristics. Future studies will address this issue by enhancing the NEP network, enabling it to perform mesh-size independent predictions while minimizing the computational cost. Furthermore, the trade-off between accuracy and dataset generation time should be further investigated. Lastly, future research will focus on exploring the capabilities of DeepFEA in CFD and FSI simulations, as well as in real-life application scenarios.

## CRedit authorship contribution statement

**Georgios Triantafyllou**: Conceptualization, Data curation, Formal analysis, Investigation, Methodology, Resources, Software, Validation, Visualization, Writing-original draft. **Panagiotis G. Kalozoumis**: Conceptualization, Data curation, Formal analysis, Investigation, Methodology, Resources, Software, Supervision, Validation, Visualization, Writing-original draft, Writing- review & editing. **George Dimas**: Conceptualization, Formal analysis, Investigation, Methodology, Resources, Software, Supervision, Validation, Writing-original draft, Writing- review & editing. **Dimitris K. Iakovidis**: Conceptualization, Formal analysis, Funding acquisition, Investigation, Methodology, Project administration, Resources, Supervision, Validation, Writing- review & editing.

## Declaration of competing interest

The authors declare that they have no known competing financial interests or personal relationships that could have appeared to influence the work reported in this paper.

## Acknowledgements

This work has been funded by the European Union, under grant agreement No 101099145, project SoftReach, (https://softreach.eu/).

## Data Availability

The datasets used in this paper can be found at https://doi.org/10.5281/zenodo.10870936.